\documentclass[runningheads]{llncs}
\usepackage{capt-of}
\usepackage{graphicx}%
\usepackage{multirow}%
\usepackage{mathrsfs}%
\usepackage{xcolor}%
\usepackage{textcomp}%
\usepackage{manyfoot}%
\usepackage{booktabs}%
\usepackage{algorithm}%
\usepackage{algorithmicx}%
\usepackage{algpseudocode}%
\usepackage{listings}%
\usepackage{diagbox}
\usepackage{float}
\usepackage{xcolor} 
\usepackage{graphicx}
\usepackage{circuitikz}
\usepackage{hhline}
\usepackage{slashbox}

\begin{document}

\title{Sim2Real in endoscopy segmentation with a novel structure aware image translation}
\titlerunning{Sim2Real in endoscopy segmentation}
\author{Clara Tomasini \and Luis Riazuelo \and Ana C. Murillo}
\institute{DIIS - I3A - Universidad de Zaragoza, Zaragoza, Spain\\
\email{ctomasini@unizar.es}}

\maketitle

\begin{abstract}
Automatic segmentation of anatomical landmarks in endoscopic images can provide assistance to doctors and surgeons for diagnosis, treatments or medical training. However, obtaining the annotations required to train commonly used supervised learning methods is a tedious and difficult task, in particular for real images. While ground truth annotations are easier to obtain  for synthetic data, models trained on such data often do not generalize well to real data. 
Generative approaches can add realistic texture to it, but face difficulties to maintain the structure of the original scene. 
The main contribution in this work is a novel image translation model that adds realistic texture to simulated endoscopic images while keeping the key scene layout information. Our approach produces realistic images in different endoscopy scenarios. We  demonstrate these images can effectively be used to successfully train a model for a challenging end task without any real labeled data. 
In particular, we demonstrate our approach for the task of fold segmentation in colonoscopy images. Folds are key anatomical landmarks that can occlude parts of the colon mucosa and possible polyps. 
Our approach generates realistic images maintaining the shape and location of the original folds, after the image-style-translation, better than existing methods.
We run experiments both on a novel simulated dataset for fold segmentation, and real data from the EndoMapper (EM) dataset~\cite{azagra2023endomapper}. All our new generated data and new EM metadata is being released to facilitate further research, as no public benchmark is currently available for the task of fold segmentation.
\end{abstract}

\section{Introduction}\label{intro}
Automatic detection of anatomical landmarks in endoscopic images can provide assistance to doctors and surgeons for diagnosis, treatments or medical training.
To accurately detect key landmarks and provide good quality information to doctors, most models require some labeled data for training. However, labeling real endoscopic images is  difficult due to the cost and expertise required. 
Simulated 3D models or real CT scans can be used to automatically obtain labels for certain landmarks that can be identified analyzing discontinuities in the surface, such as polyps or folds. 
However, while anatomically correct and automatically labeled, this type of generated images lack realistic texture, hindering the generalization to real images. 

\begin{figure}[!tb]
    \centering
    \includegraphics[width=0.9\linewidth]{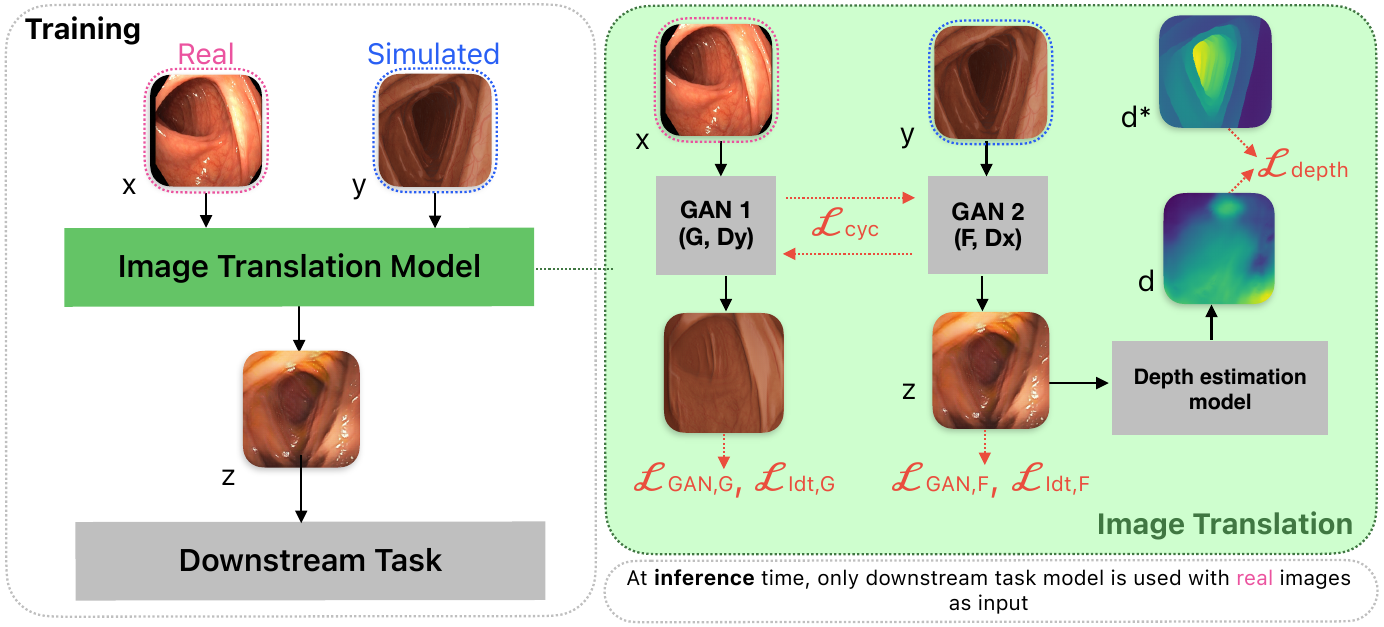}
    \caption{Overview of the proposed \textbf{new translation approach} and subsequent downstream task. During \textbf{training}, our pipeline uses {\color{magenta}\textbf{real}} (x) and {\color{blue}\textbf{synthetic}} (y) data to learn a model for \textbf{image translation}, to obtain synthetic data with realistic texture (z) while maintaining the most relevant layout structure through depth consistency. The downstream task model can then be trained supervised using the generated data. During \textbf{inference}, the model previously trained on our generated data is applied directly to {\color{magenta}\textbf{real}} images.}
    \label{fig:approach}
\end{figure}

Realistic texture can be added by means of numerous  generative models for image translation, or image style-transfer, which can apply a target texture to a source image. However, existing approaches tend to modify important parts of the original structure. 
Our work improves existing image translation methods  with a explicit constraint to maintain key scene layout information. This allows us to use the  ground-truth labels automatically obtained from the 3D models. 
The presented pipeline (see Figure~\ref{fig:approach}) is evaluated on the particular task of fold segmentation, for which no public labeled data is available. In Optical Colonoscopy (OC), commonly used to screen for colorectal cancer, folds are protrusions on the colon wall that can hide parts of the colon and cause a significant number of undetected polyps~\cite{Pickhardt2004LocationOA}, resulting in wrong diagnosis. Segmenting them can provide real-time guidance on important areas left to explore, relevant, for instance, to train new doctors. We leverage simulated data as the only supervision for a fold segmentation model. Due to the difficulty in obtaining reliable annotations and the lack of public annotated data, we build a benchmark from VR-CAPS synthetic data~\cite{incetan2020vrcaps} that can be automatically labeled. In summary, the  contributions in this work are:
\begin{itemize}
    \item A \textbf{new image translation approach} able to keep the original image structure, demonstrated on several endoscopic scenarios. 
    \item A \textbf{demonstration} of our proposed approach for the end task of \textbf{fold segmentation in colonoscopy images}. It shows both the quality of the translation step and the significant improvement on fold segmentation quality with respect to existing approaches.
    \item A \textbf{novel benchmark}\footnote{Code  and annotations available at \url{https://github.com/ropertUZ/Sim2Real-EndoscopySegmentation}} for fold segmentation composed of automatically annotated synthetic data, with added realistic texture, and manually labeled data from real colonoscopy recordings.
\end{itemize}

\section{Related Work}\label{related}

\textbf{Synthetic Data Supervision.} Synthetic data is often used as supervision in endoscopy for various tasks, for example segmentation of instruments in colonoscopy~\cite{sahu2020endo} or organs in laparoscopy~\cite{pfeiffer2019generating}. It is directly used to train supervised models when labeled real data is scarce. One end task with such scenario, and therefore with solutions proposed trained on synthetic data, is fold segmentation in real colonoscopy images. Most existing fold segmentation models use synthetic, virtual colonoscopies (VC) rather than real, optical colonoscopies (OC) and rely on direct mathematical methods like mesh gradients~\cite{zhu2012haustral} or geometric properties of the curvature of the folds~\cite{zhu2011haustral}. While these methods perform well on VC images, they cannot be directly applied to OC images. 
One recently proposed model~\cite{jin2023self} relies solely on OC images but only performs edge detection. The only model that attempts complete fold segmentation in OC images is FoldIt~\cite{mathew2021foldit}. Trained on both OC and VC images, it is based on CycleGAN~\cite{zhu2017unpaired}, an unsupervised generative model designed to perform translation between various unpaired image domains. It was evaluated on simulated endoscopic images for which ground-truth labels are not currently public, and it was not quantitatively evaluated on real endoscopic images. 

\textbf{Image Translation.} Synthetic data often lacks realistic appearance, and requires additional image translation strategies to make it look realistic and useful for training. To bridge the gap between simulated and real data, Endo-Sim2Real~\cite{sahu2020endo} trains the target model (e.g. segmentation model) directly on both domains via consistency (unlabeled real images) and supervised (labeled synthetic images) learning. Similarly, SPIGAN~\cite{lee2018spigan} simultaneously trains the target model along with a CycleGAN model~\cite{zhu2017unpaired} for transferring the realistic texture to the synthetic image and a depth estimation model to maintain the key scene layout in the image. Another strategy is to use a two-step approach, first image translation to add realistic texture to the synthetic images then train the model needed for the desired downstream task. This way, the realistic synthetic images can be used for various different tasks instead of just one, as done with the one-step approaches previously mentioned. For example, I2I~\cite{pfeiffer2019generating} uses a modified MUNIT framework~\cite{huang2018multimodal} for image translation with added MS-SSIM loss to ensure content consistency, then trains a segmentation model on these images.
The image translation step can also be done using style transfer models, such as AdaIN~\cite{huang2017arbitrary}, which consider two images, a style image and a content image, and aim at transferring texture information between them without modifying the key information of the content image. More recent approaches use transformer-based models, like StyTR2~\cite{deng2022stytr2}, or diffusion-based models, like InST~\cite{zhang2023inversion}. 
A downside to these models is the amount of data they require for training. For example, authors of StyTR2~\cite{deng2022stytr2} report training using around 130000 images and 4 GPUs. On the other hand, CycleGAN authors report using around 5000 images for most experiments and one GPU.
We consider approaches CycleGAN, I2I, AdaIN and StyTR2 in our experiments.

\textbf{Foundation Models.} To take advantage of existing models and information during training, a recurrent strategy nowadays is to incorporate the use of foundation models into different downstream task solutions. As more general purpose tools pre-trained on a large amount of data, foundation models are able to capture relevant general features from the input, and can then be used for various downstream tasks, such as segmentation tasks leveraging SAM~\cite{kirillov2023segment} or monocular depth estimation tasks with Depth Anything~\cite{depthanything}.  More recently, EndoFM~\cite{wang2023foundation}, an endoscopy-specific foundation model based on the Vision Transformer model, was released, after being trained on numerous endoscopy datasets. It shows improved results for several tasks in this domain when used as backbone within state-of-the-art models like TransUnet~\cite{chen2021transunet}. 

\section{Methodology}
Our approach for image translation relies on a CycleGAN-based approach improved with  depth consistency supervision. We generate simulated data and aim to use it to train a supervised model. We focus on the specific task of fold segmentation in colonoscopy. Figure~\ref{fig:approach} illustrates the proposed pipeline, which consists of three steps: data generation, image translation and segmentation as an end-task supervised with the \textit{translated} images.\\

\begin{figure}[!b]
    \centering
    \setlength\tabcolsep{1.5pt}
    \begin{tabular}{c|c}
         \begin{tabular}{ccc}
             \includegraphics[width=0.12\linewidth]{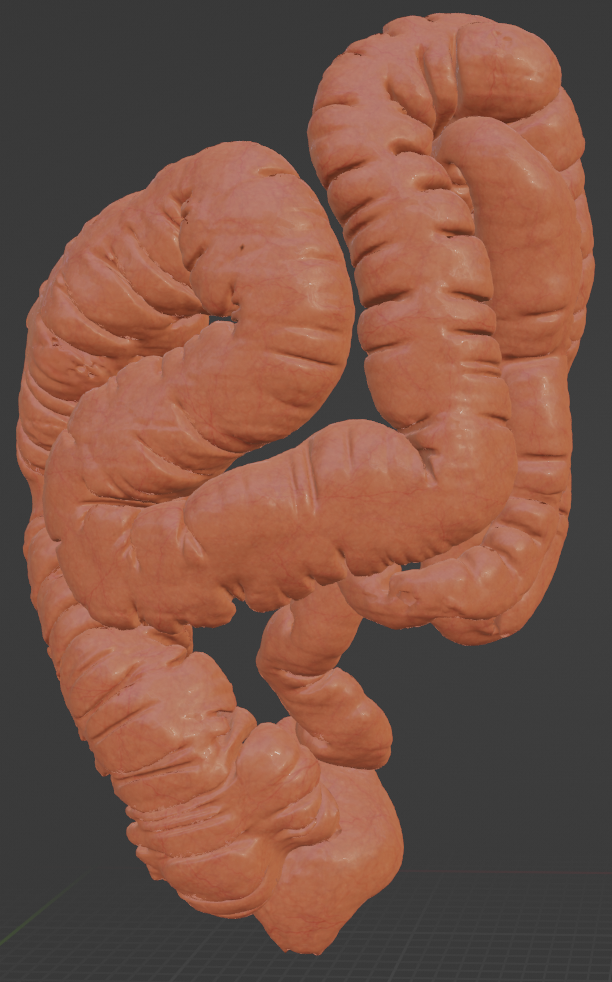} & \includegraphics[width=0.12\linewidth]{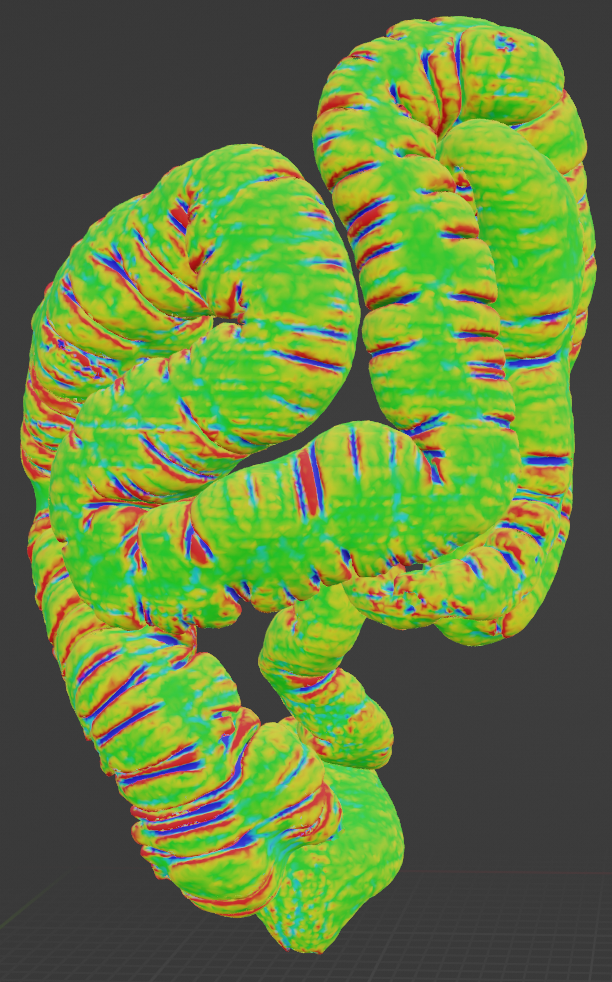}&
             \includegraphics[width=0.12\linewidth]{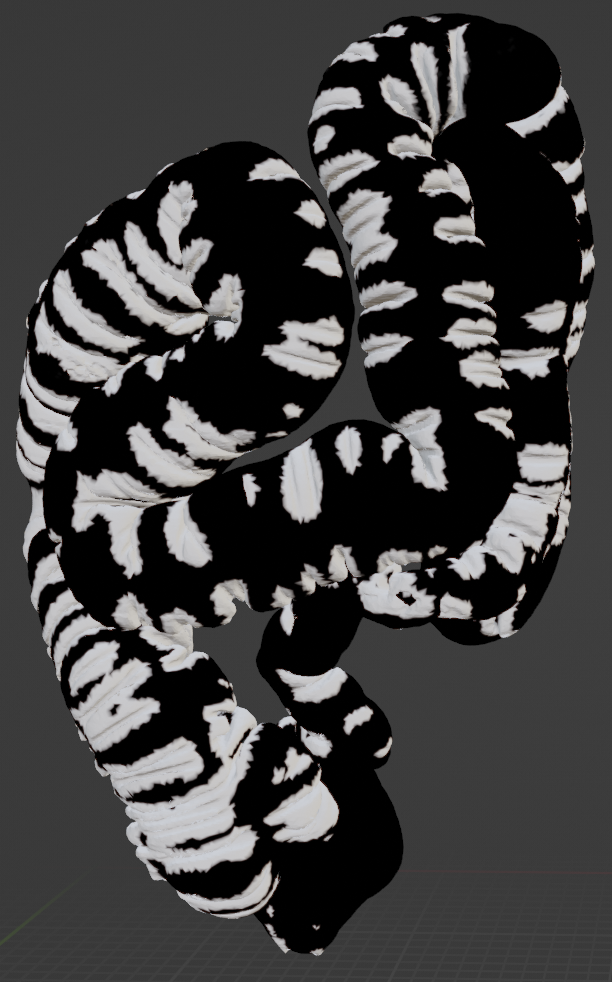} \\
             (a) & (b) & (c)\\
         \end{tabular}
         &
         \begin{tabular}{cccc}   
            \includegraphics[width=0.09\linewidth]{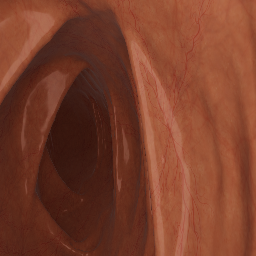} &             
            \includegraphics[width=0.09\linewidth]{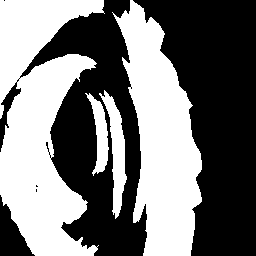} &
            \includegraphics[width=0.09\linewidth]{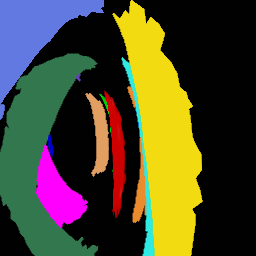} &
            \includegraphics[width=0.09\linewidth]{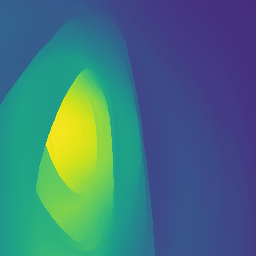} \\
            \includegraphics[width=0.09\linewidth]{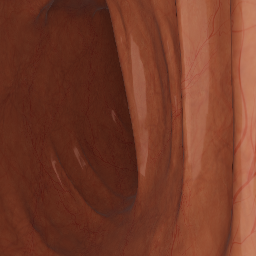} &             
            \includegraphics[width=0.09\linewidth]{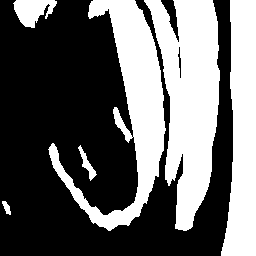} &
            \includegraphics[width=0.09\linewidth]{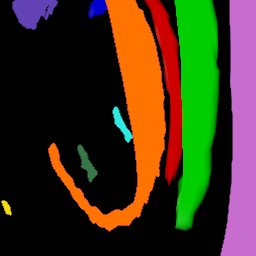} &
            \includegraphics[width=0.09\linewidth]{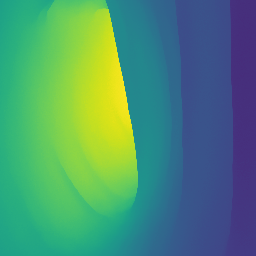} \\
            (d) & (e) & (f) & (g) \\
         \end{tabular}
         \\  
    \end{tabular}
    \caption{Labeled data generation. (a) Simulated colon from VR-CAPS, (b) curvature principal directions, and (c) binary segmentation. (d) Simulated images from VR-CAPS with corresponding (e) binary masks (f) instance masks and (g) depth maps.}
    \label{fig:vrcaps}
\end{figure}

\textbf{Data Generation and Automatic Labeling.}\label{data}
Given the difficulty in obtaining ground-truth fold segmentations, we build a new  benchmark from simulated images from VR-CAPS~\cite{incetan2020vrcaps}, and automatically label them  based on mesh geometry. A 3D mesh of the colon is obtained from CT scans, through which we define a trajectory. Images are generated at frame rate 50fps along that trajectory.
Ground-truth binary and instance fold segmentations are obtained by automatically computing the curvature principal directions in the mesh. The folds correspond to areas where that direction changes. 
We also generate ground-truth depth maps. Figure~\ref{fig:vrcaps}(a)-(c) shows the simulated colon with VR-CAPS texture, the automatically computed curvature direction changes and consequent binary segmentation. Figure~\ref{fig:vrcaps}(d)-(g) shows examples of frames with binary and instance segmentations and depth maps generated with our approach.\\  

\textbf{Image Translation.}\label{translation}
Our novel image translation model is built on CycleGAN. CycleGAN provides one of the best style transfer solutions, but, as shown in Figure~\ref{fig:translation}, it often outputs an image with a significantly modified layout compared to the original input structure. This hinders the use of labels obtained in the original input data. 
To alleviate this issue, we propose to encourage depth consistency, and therefore maintain scene structural content, during training. 
We add a new depth consistency loss term, eq.~(\ref{eq:loss}), to the overall CycleGAN loss via the scale-invariant depth log loss function~\cite{eigen2015predicting}. We consider the simulated input image $x$ from domain X (simulated), and its corresponding ground-truth depth map $d^*$. Generator G outputs the corresponding image $y$ in domain Y (simulated with realistic texture). 
A pretrained model, fixed during training of CycleGAN, predicts a depth map $d$ for image $y$. The loss is calculated over the $n$ pixels of each image as 
\begin{equation}
    \displaystyle \mathcal{L}_{depth} \ =\ \frac{1}{n}\sum _{i} (log(d_{i}) - log(d^*_{i}))^{2} \ -\frac{1}{2n^{2}}\left(\sum _{i} (log(d_{i}) - log(d^*_{i}))\right)^{2}. \
    \label{eq:loss}
\end{equation}

\textbf{Downstream Segmentation task.}\label{segmentation}
For the segmentation end task, we build a model based on TransUNet~\cite{chen2021transunet} but including the EndoFM backbone~\cite{wang2023foundation}, which is available pretrained on seven different real endoscopy datasets. We train the segmentation model fully supervised on our simulated data after applying the style transfer. During training, we consider our two versions of each simulated image. In particular, we build the training batches using pairs of images with original texture and transferred realistic texture, to mitigate the slight distortion still introduced by the translation model.

\section{Experimental results}
\subsection{Image Translation in different endoscopy scenarios.}
\begin{figure}[!t]
    \scriptsize
    \centering
    \setlength\tabcolsep{1.5pt}
    \begin{tabular}{cc@{\hspace{4.5mm}}cccc@{\hspace{4.5mm}}c}
          Target & Original & CycleGAN & I2I & $StyTR^2$ & AdaIN & Ours\\
         Style & image
         & \cite{zhu2017unpaired} & \cite{pfeiffer2019generating} & \cite{deng2022stytr2} & \cite{huang2017arbitrary} & \\

        \includegraphics[width=0.1\linewidth]{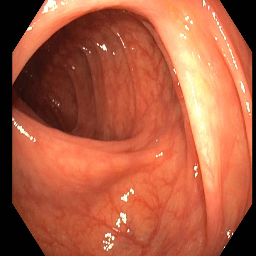}&
        \includegraphics[width=0.1\linewidth]{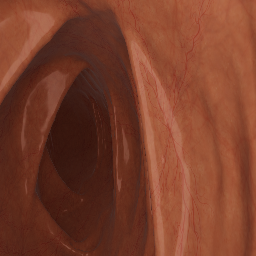} & \includegraphics[width=0.1\linewidth]{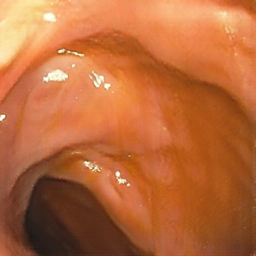} &
        \includegraphics[width=0.1\linewidth]{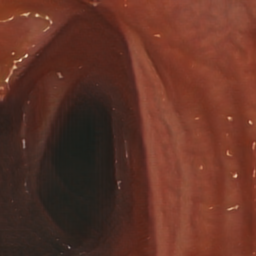} & 
        \includegraphics[width=0.1\linewidth]{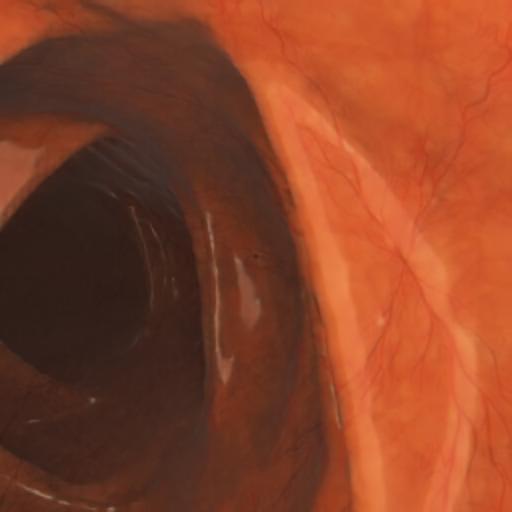} & 
        \includegraphics[width=0.1\linewidth]{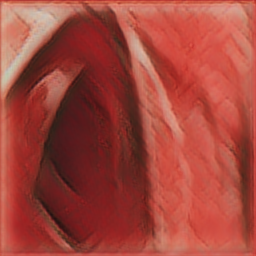} &
        \includegraphics[width=0.1\linewidth]{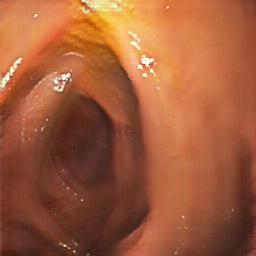}
        \\
        
        \includegraphics[width=0.1\linewidth]{images/image_translation/real/seq_13_frame_1021.png}&
        \includegraphics[width=0.1\linewidth]{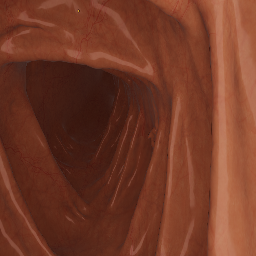}& 
        \includegraphics[width=0.1\linewidth]{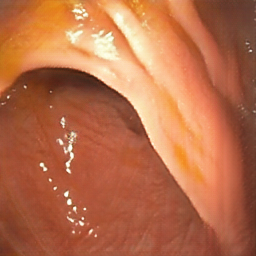} & 
        \includegraphics[width=0.1\linewidth]{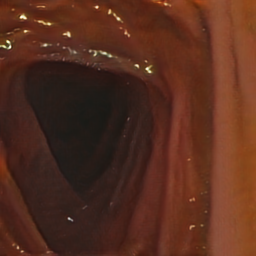} & 
        \includegraphics[width=0.1\linewidth]{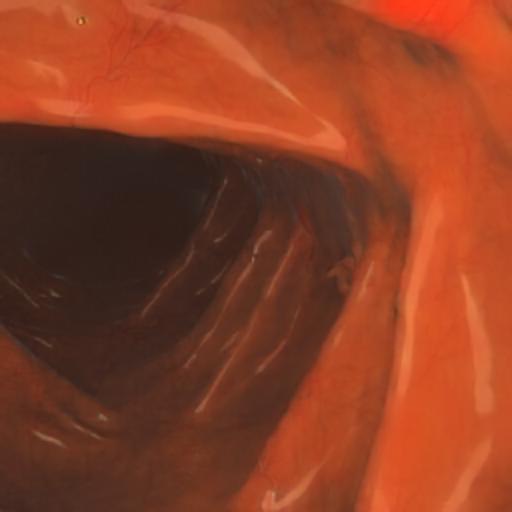} & 
        \includegraphics[width=0.1\linewidth]{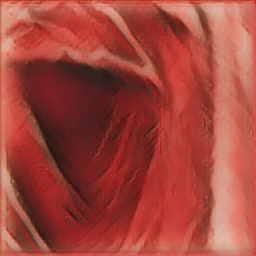} &
        \includegraphics[width=0.1\linewidth]{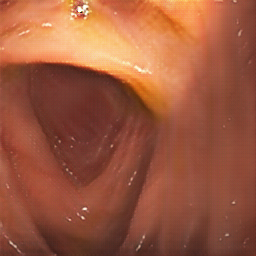}     
    \end{tabular}
    \caption{Simulated \textbf{colonoscopy} images processed with different style transfer models to have realistic texture. CycleGAN, I2I and Ours are trained on 1400 simulated images from VR-CAPS 
    and 1400 real images from EM datasets. 
    $StyTR^2$ and AdaIN models are used as provided, pretrained, and only take one target style input example. SCSfMLearner~\cite{bian2021unsupervised} is used pretrained on VR-CAPS~\cite{incetan2020vrcaps} for depth inference in our approach.  
    }
    \label{fig:translation}
\end{figure}

To illustrate our approach and  its advantages, we compare  various style transfer models to put the texture of real colonoscopy images from the EM dataset~\cite{azagra2023endomapper} to our simulated colonoscopy images. As discussed in Section~\ref{related}, we compare CycleGAN~\cite{zhu2017unpaired}, I2I~\cite{pfeiffer2019generating}, $StyTR^2$~\cite{deng2022stytr2} and AdaIN~\cite{huang2017arbitrary}. 
Figure~\ref{fig:translation} shows the resulting image for a given input using each of the methods compared. Additional examples obtained using our approach can be seen in Appendix 1 (\ref{appendixA}).
I2I, AdaIN and $StyTR^2$, while not distorting the shape of the folds, are not able to capture well the texture of the real images.  CycleGAN~\cite{zhu2017unpaired} best captures the texture, but significantly distorts the structure of the images, meaning the semantic segmentation masks available for the input simulated images will not correspond to the final images after the realistic texture is applied. 

\begin{figure}[!tb]
    \centering
    \begin{tabular}{ccc}
    \includegraphics[width=0.31\linewidth]{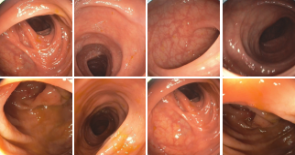} &
    \includegraphics[width=0.31\linewidth]{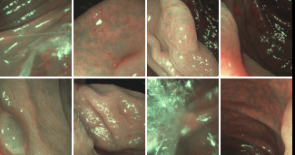} &
    \includegraphics[width=0.31\linewidth]{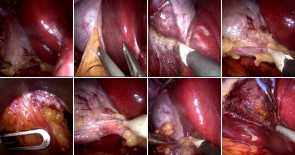}\\ 
    (a) & (b) & (c)%
    \end{tabular}
    \caption{Examples from three different datasets used as source style in our experiments. (a) Colonoscopy images from EM~\cite{azagra2023endomapper}. (b) NBI colonoscopy images from EM~\cite{azagra2023endomapper}. (c) Laparoscopy images from Cholec80~\cite{twinanda2016endonet}.}
    \label{fig:image_domains}
\end{figure}

\begin{figure}[!tb]
    \scriptsize
    \centering
    \setlength\tabcolsep{0.5pt}
    \begin{tabular}{ccccc @{\hspace{3mm}} cccccc}
          Target & Original & Cycle- & I2I & Ours & 
          Target & Original & Cycle- & I2I* & I2I & Ours \\
         Style & image & GAN~\cite{zhu2017unpaired} & \cite{pfeiffer2019generating} & &
         Style & image & GAN~\cite{zhu2017unpaired} & & \cite{pfeiffer2019generating} & \\

        \includegraphics[width=0.08\linewidth]{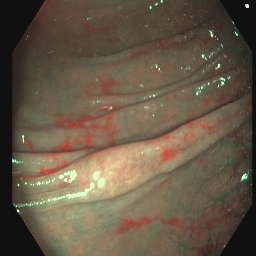}&
        \includegraphics[width=0.08\linewidth]{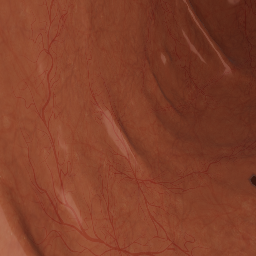}& 
        \includegraphics[width=0.08\linewidth]{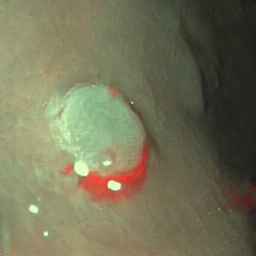} & 
        \includegraphics[width=0.08\linewidth]{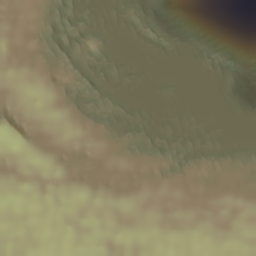} & 
        \includegraphics[width=0.08\linewidth]{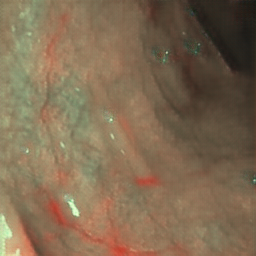} &

        \includegraphics[width=0.08\linewidth]{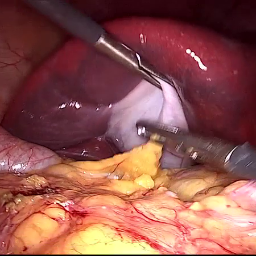}&
        \includegraphics[width=0.08\linewidth]{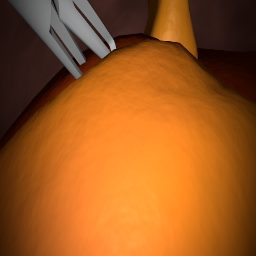}&
        \includegraphics[width=0.08\linewidth]{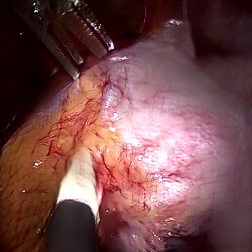}&
        \includegraphics[width=0.08\linewidth]{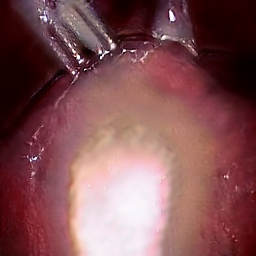}&
        \includegraphics[width=0.08\linewidth]{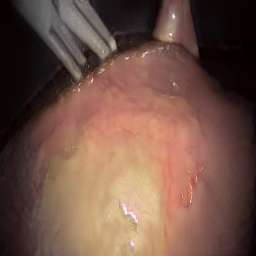}&
        \includegraphics[width=0.08\linewidth]{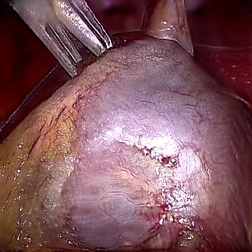}
        \\
        
        \includegraphics[width=0.08\linewidth]{images/image_translation/nbi/Seq_025_1_0134.png}& 
        \includegraphics[width=0.08\linewidth]{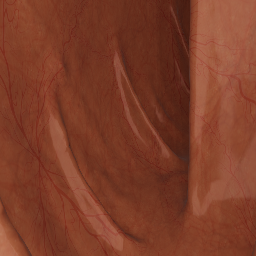} &
        \includegraphics[width=0.08\linewidth]{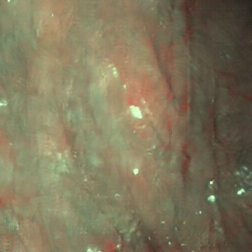} & 
        \includegraphics[width=0.08\linewidth]{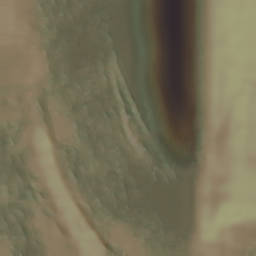} & 
        \includegraphics[width=0.08\linewidth]{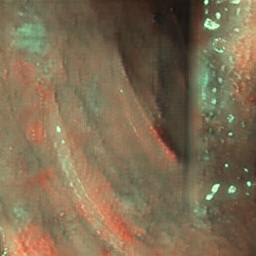} &

        \includegraphics[width=0.08\linewidth]{images/image_translation/cholec/000365.png}&
        \includegraphics[width=0.08\linewidth]{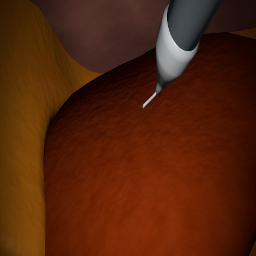}&
        \includegraphics[width=0.08\linewidth]{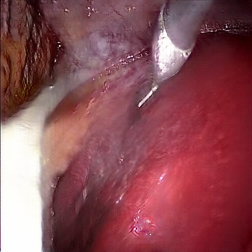}&
        \includegraphics[width=0.08\linewidth]{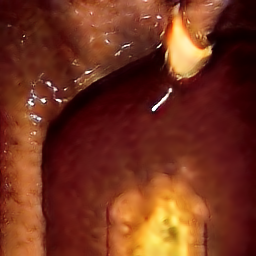}&
        \includegraphics[width=0.08\linewidth]{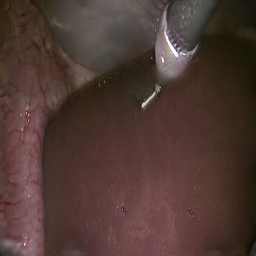}&
        \includegraphics[width=0.08\linewidth]{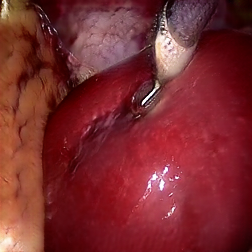}\\
        \multicolumn{5}{c}{(a) NBI style to simulated colonoscopy} & \multicolumn{6}{c}{(b) Real laparoscopy style to synthetic}
    \end{tabular}
    \caption{Style transfer using different approaches. (a) All models trained on 1400 simulated \textbf{colonoscopy} images from VR-CAPS and 3640 real \textbf{NBI colonoscopy} images from EM
    . (b) CycleGAN, I2I* and Ours trained on 2000 simulated \textbf{laparoscopic} images~\cite{pfeiffer2019generating} and 2840 real \textbf{laparoscopic} images from Cholec80~\cite{twinanda2016endonet}. I2I corresponds to results published in~\cite{pfeiffer2019generating} trained on 20000 simulated images~\cite{pfeiffer2019generating} and 74 000 real images from Cholec80~\cite{twinanda2016endonet}. In both experiments, DepthAnything~\cite{depthanything} is used for depth inference in our approach.
    }
    \label{fig:translation_otro}
\end{figure}

We demonstrate our approach in two additional style transfer endoscopy scenarios, shown in Fig.~\ref{fig:image_domains} (b,c). The results of applying these styles, NBI light colonoscopy and real laparoscopic images, are illustrated in  Fig~\ref{fig:translation_otro}. In both cases, we observe significant structure distortion introduced by CycleGAN. Our approach results look the best for colonoscopy images under NBI lighting conditions, and look comparable to I2I in laparoscopic images, but using much less training samples. 
Note that the best results for I2I are obtained training with much more data than our approach. \\

\subsection{Fold segmentation results training only with simulated data.} 
Besides the previous qualitative results of the style transferred achieved by our approach, this section demonstrates, quantitatively, the benefits of using our style transfer approach on the end-task of fold segmentation. 

These experiments consider five different \textbf{training sets} and three \textbf{evaluation sets}, as summarized in Table \ref{tab:datasets}. We measure the quality of the \textbf{binary segmentations} obtained with the commonly used Intersection over Union (IoU) metric on the available annotated frames. 
\begin{table}[!b]
    \centering
    \caption{Datasets used for training and evaluating our proposed pipeline: name of the dataset, phase in which it is used (training or testing),  number of frames in the dataset, additional information (source, available annotations).}
    \label{tab:datasets}
    \begin{tabular}{c|c|c|c}
        \textbf{Name} & \textbf{Step} & \textbf{\# Frames} & \textbf{Dataset details} \\
        \hline
        \hline
        \multirow{2}{*}{\textbf{EM}} & Train & 1400 (Train) & Real images from EndoMapper dataset~\cite{azagra2023endomapper}. \\ 
         & Test & 100 (Test) & Test set contains manual annotations.  \\
        \hline
        \multirow{2}{*}{\textbf{Sim.}} & Train & 1400 (Train) & Simulated images from VR-CAPS~\cite{incetan2020vrcaps}\\
        & Test & 500 (Test) & Contains GT fold labels and depth maps. \\
        \hline
        \multirow{2}{*}{\textbf{Sim-Aug}.} & Train & 1400 (Train) & Sim. set with added realistic texture. \\
         & Test & 500 (Test) & \\
        \hline
        \textbf{FI-OC} & Train & 1800 & Real images from HyperKvasir dataset~\cite{borgli2020hyperkvasir}. \\
        & & & Used to train FoldIt~\cite{mathew2021foldit}.\\
        \hline
        \textbf{FI-VC} & Train & 1800 & VC images from CT Colonography TCIA data \cite{tcia2015}.\\
        & & & Used to train FoldIt~\cite{mathew2021foldit}. Contains GT fold labels.

    \end{tabular}
\end{table}

All the training and evaluation is run on a GPU GeForce RTX3090 on a desktop. 
EndoFM-TransUNet model is trained for 150 epochs with learning rate $1e-2$. FoldIt model is trained as recommended by the authors~\cite{mathew2021foldit}. Both training and testing images have size 256x256.\\ 

Table~\ref{tab:res_bin} summarizes the binary fold segmentation results obtained with our proposed pipeline and the FoldIt baseline variations. 
We show an overall 11.66\% improvement on the EM  evaluation set using our approach, compared to the best performing FoldIt baseline, of which 4.6\% is the result of our proposed image translation and paired batch training structure.
More specifically, without any translation method (training only on simulated data), EndoFM-TU provides a 7.06\% improvement compared to FoldIt trained on real data from EM. train set. Adding CycleGAN or I2I image translation to add realistic texture to the simulated images used for training EndoFM-TU does not improve the results, instead making them worse, while our approach does provide a further 2.8\%.
Finally, the proposed training batch, consisting of pairs of images with and without realistic texture (Sim. and Sim-Aug. sets) adds an additional 1.8\% performance improvement.
Figure~\ref{fig:res_img} shows segmentation examples obtained with our approach trained on the Sim. and Sim-Aug. sets and FoldIt trained on EM and FoldIt-VC sets (best results obtained by FoldIt on EM test set), in the EM test set. It qualitatively confirms the improved results of our approach with respect to the baseline. 
Our segmentation appears less noisy, with less false positives, as illustrated in the last example of the figure. Additional examples of segmentations obtained using our augmented images during training can be seen in Appendix 2 (\ref{appendixB}).

\begin{table}[!bt]
    \centering
    \setlength\tabcolsep{6pt}
    \caption{Quantitative binary segmentation results. Mean IoU obtained with variations of our proposed pipeline and the baseline model FoldIt. Training and evaluation sets include both {\color{magenta}\textbf{Real}} and {\color{blue}\textbf{Simulated}} data.}
    
   \label{tab:res_bin} 
      \begin{tabular}{c|c|c|cc}
      \toprule
       \textbf{Model}  & \textbf{Train} & 
       \textbf{Translation}&
       \multicolumn{2}{@{}c@{}}{\textbf{Evaluation Set}}\\
        & \textbf{Set} & \textbf{Method} &
        {\color{blue}\textbf{Sim-Aug~}} & {\color{magenta}\textbf{EM\cite{azagra2023endomapper}}} \\ 
       \hline
       \hline
      \multirow{7}{*}{\rotatebox[origin=c]{90}{FoldIt~\cite{mathew2021foldit}}} &   {\color{magenta}\textbf{FI-OC}} \& & & &\\
      &{\color{blue}\textbf{FI-VC}}  & N.A. & 36.50 $\pm$ 11.19 & 29.06 $\pm $10.98 \\ 
      \cline{2-5}
      &  {\color{magenta}\textbf{EM}} \& & & &\\
      & {\color{blue}\textbf{FI-VC}} & N.A. & 30.15 $\pm$ 10.73 & 32.64 $\pm$ 10.26 \\ 
      \cline{2-5}
      & {\color{blue}\textbf{Sim-Aug.}} \& & & &\\
      & {\color{blue}\textbf{Sim.}} \& & & & \\
      & {\color{blue}\textbf{FI-VC}} & Ours & 23.99 $\pm$ 9.67 & 21.07 $\pm$ 9.68 \\
     \hline
      \multirow{6}{*}{\rotatebox[origin=c]{90}{EndoFM-TU}} & {\color{blue}\textbf{Sim.}} & N.A. & 38.1 $\pm$ 8.56 & 39.7 $\pm$ 12.83 \\ 
      \cline{2-5}
       & {\color{blue}\textbf{Sim-Aug.}} & CycleGAN 
       & 40.7 $\pm$ 12.25 & 26.9 $\pm$ 9.61\\ 
       \cline{2-5}
     & {\color{blue}\textbf{Sim-Aug.}} & I2I 
       & 35.2 $\pm$ 10.73 & 37.6 $\pm$ 12.67\\
       \cline{2-5}
       & {\color{blue}\textbf{Sim-Aug.}} & Ours & \textbf{48.9} $\pm$ 9.62 & 42.5 $\pm$ 11.59\\ 
       \cline{2-5}
       & {\color{blue}\textbf{Sim-Aug.}} \& & & &\\
       &{\color{blue}\textbf{Sim.}} & Ours & 42.6 $\pm$ 10.68 & \textbf{44.3} $\pm$ 11.47\\ 
      \hline
      \end{tabular}
\end{table}

\begin{figure}[!bt]
    \centering
    \setlength\tabcolsep{1pt}
    \begin{tabular}{cccc}
         Original & GT & FoldIt & Ours\\ \includegraphics[width=0.11\linewidth]{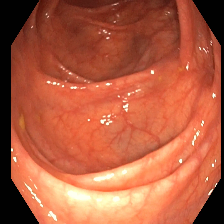}  &
         \includegraphics[width=0.11\linewidth]{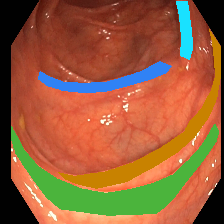}  &
         \includegraphics[width=0.11\linewidth]{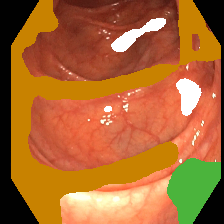}  &
         \includegraphics[width=0.11\linewidth]{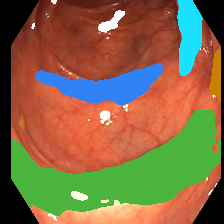}\\
        
         \includegraphics[width=0.11\linewidth]{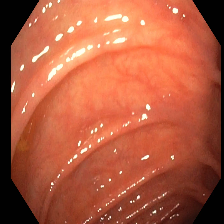}  & 
         \includegraphics[width=0.11\linewidth]{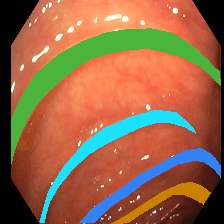}  &
         \includegraphics[width=0.11\linewidth]{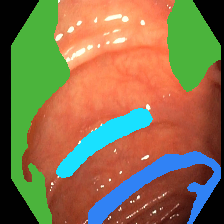}  &
         \includegraphics[width=0.11\linewidth]{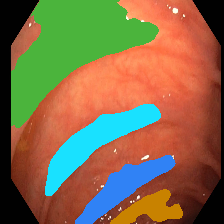}\\
         
         \includegraphics[width=0.11\linewidth]{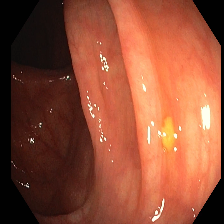}  &
         \includegraphics[width=0.11\linewidth]{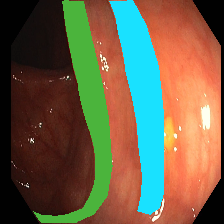} &
         \includegraphics[width=0.11\linewidth]{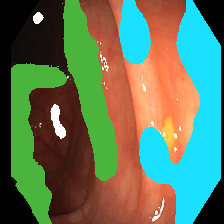} &
         \includegraphics[width=0.11\linewidth]{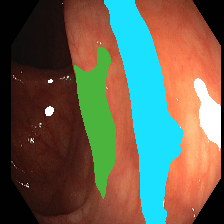} \\
    \end{tabular}
    \caption{\textbf{Segmentation results.} Each color is a different fold instance segmented. White segments are instances without sufficient overlap to any ground-truth  instance.
    FoldIt trained on EM and FoldIt-VC. Our approach trained on Sim. and Sim-Aug.}
    \label{fig:res_img}
\end{figure}

\section{Conclusion}
While obtaining labels for real images can be difficult due to cost and time limitations, they can be more easily generated for simulated images. However, simulated images usually lack realistic texture, leading to models trained exclusively on simulated data to not generalize well to real data. 
We present an improved image translation approach to add realistic texture to simulated images, enforcing a better scene layout consistency, that allows us to train downstream models relying only on labeled simulated data. We show that the resulting model is able to perform well on real data, therefore bridging the gap between real and simulated scenarios.
We evaluate the performance of our approach on the specific task of fold segmentation in colonoscopy. Folds are relevant colonoscopy landmarks, with challenging detection due to intrinsic image properties and with limited tools and labeled data availability. Our pipeline achieves significant improvement for the task, potentially leading to novel applications to assist doctors during endoscopy procedures, and could also be expanded to segmentation of other key endoscopy landmarks like polyps and lesions.  
We also build and release a simulated data fold segmentation benchmark, with ground-truth labels and depth maps, to facilitate further work on the fold segmentation problem. 

\section{\ackname} This project has been funded by the European Union’s Horizon 2020 research and innovation programme under grant agreement No 863146 and Aragón Government project T45\_23R.
\section{\discintname} The authors declare no competing interests.

\section{Appendix 1: Additional Image Translation Examples} \label{appendixA}
\begin{figure}
    \centering
    \includegraphics[width=1\linewidth]{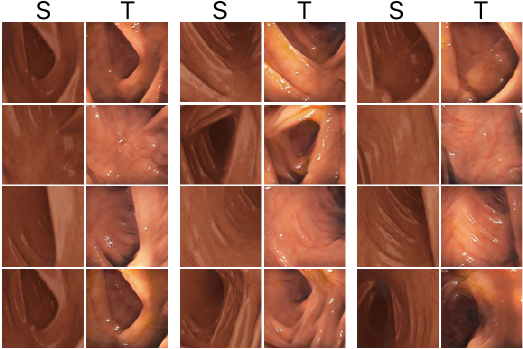}
    \caption{Examples of simulated colonoscopy images processed with our image translation approach to have realistic texture learned on the EndoMapper dataset. Images are shown by pairs of simulated image (S), and corresponding translated image (T).}
    \label{fig:translation_ex}
\end{figure}

\section{Appendix 2: Additional Fold Segmentation Examples}\label{appendixB}
\begin{figure}
\centering
        \begin{tabular}{cccc}
        Original & GT & FoldIt & Ours \\
         \includegraphics[width=0.15\linewidth]{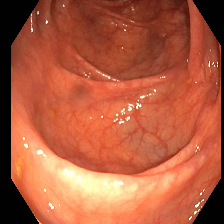} & 
        \includegraphics[width=0.15\linewidth]{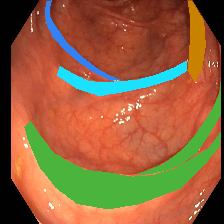}  &
        \includegraphics[width=0.15\linewidth]{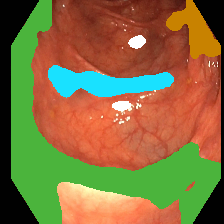}  &
        \includegraphics[width=0.15\linewidth]{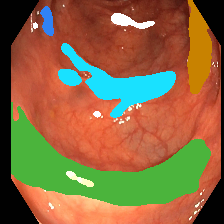} \\
         
         \includegraphics[width=0.15\linewidth]{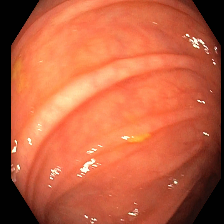}  & 
        \includegraphics[width=0.15\linewidth]{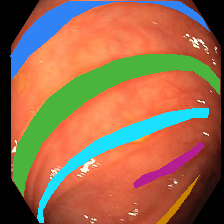}  &
        \includegraphics[width=0.15\linewidth]{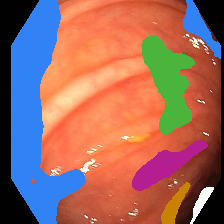}  &
        \includegraphics[width=0.15\linewidth]{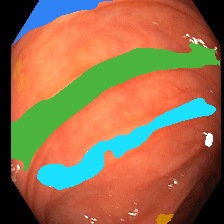} \\
         
         \includegraphics[width=0.15\linewidth]{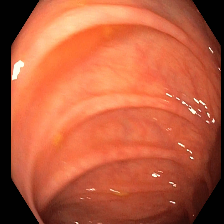}  & 
         \includegraphics[width=0.15\linewidth]{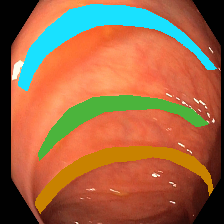}  &
         \includegraphics[width=0.15\linewidth]{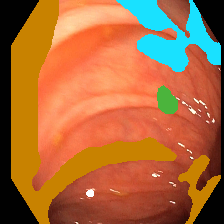}  &
          \includegraphics[width=0.15\linewidth]{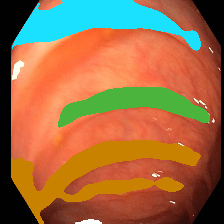} \\
          
         \includegraphics[width=0.15\linewidth]{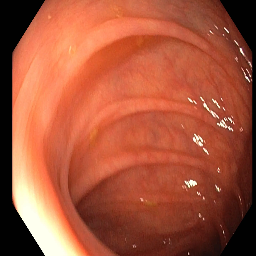}  & 
         \includegraphics[width=0.15\linewidth]{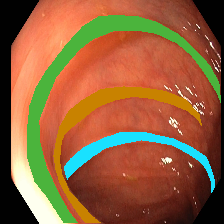}  &
         \includegraphics[width=0.15\linewidth]{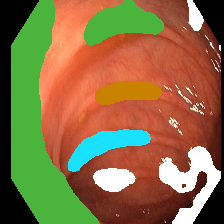}  &
         \includegraphics[width=0.15\linewidth]{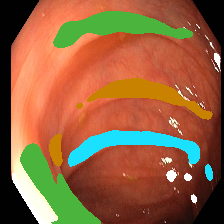} \\
         
         \includegraphics[width=0.15\linewidth]{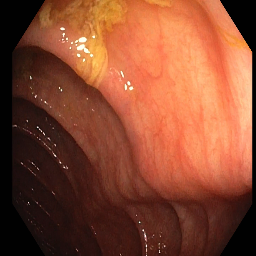}  & 
         \includegraphics[width=0.15\linewidth]{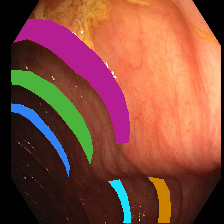}  &
         \includegraphics[width=0.15\linewidth]{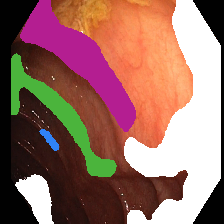}  &
         \includegraphics[width=0.15\linewidth]{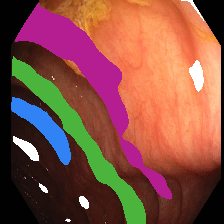} \\
         
         \includegraphics[width=0.15\linewidth]{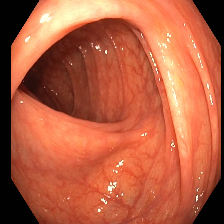} &
          \includegraphics[width=0.15\linewidth]{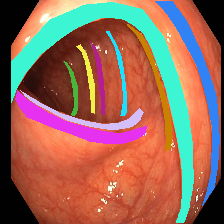} &
          \includegraphics[width=0.15\linewidth]{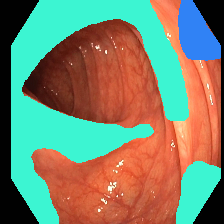}  &
         \includegraphics[width=0.15\linewidth]{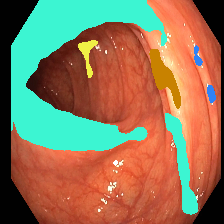} \\
        \end{tabular}
\caption{Examples of fold segmentation in real colonoscopy images. Results obtained with our approach, and the closest work in the literature (FoldIt), trained only on simulated data with realistic texture added thanks to our image translation approach. Each color indicates a different fold instance segmented. White segments 
are instances without sufficient overlap to any ground-truth (GT) instance.}
\label{fig:seg}
\end{figure}

\bibliographystyle{splncs04}
\bibliography{bibliography}

\end{document}